\documentclass{article}
\usepackage[backend=biber,style=numeric,sorting=nyt]{biblatex} 
\usepackage{booktabs}
\usepackage{colortbl}
\usepackage{longtable}
\newtheorem{researchq}{Research Question}
\usepackage{arxiv}
\usepackage[utf8]{inputenc} 
\usepackage[T1]{fontenc}    
\usepackage{hyperref}       
\usepackage{url}            
\usepackage{booktabs}       
\usepackage{amsfonts}       
\usepackage{nicefrac}       
\usepackage{microtype}      
\usepackage{lipsum}		
\usepackage{graphicx}
\usepackage{tabularx}
\usepackage{doi}
\usepackage{fancyhdr}
\usepackage{multirow}
\bibliography{Research Report Literature}
\title{Useful for Exploration, Risky for Precision: Evaluating AI Tools in Academic Research}

\author{ 
    \href{https://orcid.org/0009-0000-2295-6976}{\includegraphics[scale=0.06]{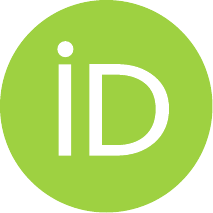}\hspace{1mm} Anthea ~Dathe} \\
	Department Speculative Transformation\\
	Dresden University of Technology\\
	Dresden, 01062 \\
	\texttt{anthea.dathe@mailbox.tu-dresden.de} \\
	\And
	\href{https://orcid.org/0009-0003-5329-599X}{\includegraphics[scale=0.06]{orcid.pdf}\hspace{1mm} Kiran ~Hoffmann} \\
	Department Speculative Transformation\\
	Dresden University of Technology\\
	Dresden, 01062 \\
	\texttt{kiran.hoffmann@mailbox.tu-dresden.de} \\
    \And
    \href{https://orcid.org/0009-0006-8240-0922}{\includegraphics[scale=0.06]{orcid.pdf}\hspace{1mm}Aline ~Mangold} \\
	Department Speculative Transformation\\
	Dresden University of Technology\\
	Dresden, 01062 \\
	\texttt{aline.mangold@tu-dresden.de} \\
    }

\begin{document}
\maketitle

\begin{abstract}
Artificial intelligence (AI) tools are being incorporated into scientific research workflows with the potential to enhance efficiency in tasks such as document analysis, question answering (Q\&A), and literature search. However, system outputs are often difficult to verify, lack transparency in their generation and remain prone to errors. Suitable benchmarks are needed to document and evaluate arising issues. Nevertheless, existing benchmarking approaches are not adequately capturing human-centered criteria such as usability, interpretability, and integration into research workflows. To address this gap, the present work proposes and applies a benchmarking framework combining human-centered and computer-centered metrics to evaluate AI-based Q\&A and literature review tools for research use. The findings suggest that Q\&A tools can offer valuable overviews and generally accurate summaries; however, they are not always reliable for precise information extraction. Explainable AI (xAI) accuracy was particularly low, meaning highlighted source passages frequently failed to correspond to generated answers. This shifted the burden of validation back onto the researcher. Literature review tools supported exploratory searches but showed low reproducibility, limited transparency regarding chosen sources and databases, and inconsistent source quality, making them unsuitable for systematic reviews. A comparison of these tool groups reveals a similar pattern: while AI tools can enhance efficiency in the early stages of the research workflow and shallow tasks, their outputs still require human verification. The findings underscore the importance of explainability features to enhance transparency, verification efficiency and careful integration of AI tools into researchers’ workflows. Further, human-centered evaluation remains an important concern to ensure practical applicability.   
\end{abstract}

\section{Introduction}
Artificial intelligence (AI) is increasingly embedded in scientific research workflows, supporting tasks such as literature search, document analysis and scientific writing \cite{vannoordenAIScienceWhat2023a}. While AI tools promise substantial efficiency gains, they also introduce a fundamental tension: researchers are increasingly confronted with AI-generated outputs that are often opaque in their reasoning \cite{kamathExplainableArtificialIntelligence2021} and potentially error-prone \cite{alansariLargeLanguageModels2026}. As a result, researchers may be required to manually check or reproduce AI-generated steps to ensure accuracy and scientific validity, reducing some of the expected efficiency benefits.
This issue can be addressed by Explainable Artificial Intelligence (xAI), which aims to make the internal behavior of AI tools more accessible and understandable to human users. xAI systems provide insight into how outputs are generated, what factors influence predictions, and where model limitations lie \cite{gunningDARPAsExplainableArtificial2019}. This can increase user trust and support safer use of AI tools in scientific contexts \cite{kamathExplainableArtificialIntelligence2021}. Consequently, explainability is not merely a technical feature but a fundamental requirement for human-centered AI design, ensuring that systems are aligned with users’ cognitive needs and decision-making processes.
Within academic research, AI systems are particularly relevant in document-based question answering (Q\&A) and literature review support. In this context, we conducted two separate benchmarks of Q\&A tools and literature review tools. Q\&A tools enable natural-language interaction with scientific documents by retrieving and synthesizing relevant information from one or multiple sources, often using retrieval-augmented generation. Literature review tools, in contrast, support the discovery and exploration of relevant publications through natural-language queries and integrated search and summarization functionalities. In our study, we define literature review tools as general AI assistants that offer an unsystematic literature review functionality.
Despite rapid adoption, existing research highlights persistent limitations in current systems, including hallucinations, inconsistent retrieval behavior, weak citation reliability and limited transparency. At the same time, research suggest that usability, interpretability and workflow integration are critical factors for effective use, yet are still insufficiently captured in current benchmarking approaches. As a result, there remains a lack of integrated evaluation frameworks that combine technical performance with user-centered criteria. To address this gap, this study proposes a human-centered benchmark for AI tools in scientific research, focusing on Q\&A- and literature review tools. The benchmark evaluates tool performance across task-oriented and usability-related dimensions, aiming to provide a more comprehensive understanding of their suitability for academic workflows.

\section{Related Work}
\subsection{AI tools in academic research}
Generative Artificial Intelligence (GenAI) refers to systems that generate novel content rather than solely analyzing, retrieving, or classifying existing information \cite{bordasWhatGenerativeGenerative2024}. In scientific contexts, such systems are increasingly integrated across multiple stages of the research process, including ideation, literature exploration, synthesis, drafting, and publication preparation \cite{vannoordenAIScienceWhat2023a}. As a result, a diverse ecosystem of GenAI tools for scientific research has emerged, each targeting distinct research-related tasks. GenAI tools for scientific research can be grouped into several functional categories according to their primary functional orientation, including writing and publication support tools, document-based question and answer (Q\&A) tools, literature review tools, generalist AI assistants and visualization tools (see table \ref{tab:genai-tools} for a detailed overview). Q\&A tools can assist researchers in efficiently understanding and analyzing one or multiple research articles, for example by extracting information, answering questions or comparing findings across documents. This functionality is highly relevant, as the number of scientific publications has increased sharply in recent years. Thus, the publication-related workload per researcher has grown substantially \cite{hansonStrainScientificPublishing2024a}. At the same time, researchers face limited cognitive and time resources when navigating this expanding body of literature. Literature review tools, such as Consensus in turn, support the identification of relevant publications through natural-language queries rather than complex search strings. Some generalist AI assistants, (e.g. \textit{You.com}) offer literature review functionalities in addition to other features. For these reasons, Q\&A tools and tools providing literature review functionality are examined more closely in this research report, as they support two key stages of the research process: identifying relevant literature and analyzing scientific documents.

\newcolumntype{Y}{>{\raggedright\arraybackslash}X}

\begin{table}[htbp]
\centering
\caption{GenAI Tools and their core functionalities}
\label{tab:genai-tools}
\renewcommand{\arraystretch}{1.2}
\begin{tabularx}{\textwidth}{p{0.27\textwidth} Y p{0.22\textwidth}}
\toprule
\textbf{Category} & \textbf{Description} & \textbf{Tools} \\
\addlinespace
Writing \& Publication Support
& Polishing and drafting: writing texts, improving language, formal checks, and citation integration
& \begin{tabular}[t]{@{}l@{}}
Jenni.ai \\
Paperpal \\
Quillbot
\end{tabular} \\
\addlinespace
Literature research \& overview
& List-based exploration: primary search and orientation in scientific literature
& \begin{tabular}[t]{@{}l@{}}
Consensus \\
Semantic Scholar
\end{tabular} \\
\addlinespace
AI assistance \& knowledge synthesis
& Research co-pilot: exploration, synthesis, and communication of literature; does not replace the rigour of a systematic review and is broader in scope than a database or focused Q\&A search
& \begin{tabular}[t]{@{}l@{}}
Perplexity
\end{tabular} \\
\addlinespace
Systematic reviews \& evidence synthesis
& Structured, methodical literature research
& \begin{tabular}[t]{@{}l@{}}
Elicit
\end{tabular} \\
\addlinespace
Document-based Q\&A tools
& Analysis of your own PDFs and other document formats, Q\&A, and summaries
& \begin{tabular}[t]{@{}l@{}}
Humata \\
ScienceOS \\
PDF.ai \\
AskYourPDF \\
ChatPDF.com
\end{tabular} \\
\addlinespace
Generalist AI assistants
& Wide-ranging use of AI, including research, writing, coding, creative work, and web searches
& \begin{tabular}[t]{@{}l@{}}
ChatGPT \\
Claude \\
DeepSeek \\
You.com
\end{tabular} \\
\addlinespace
Visualization tools
& Graph-based exploration: interactive mapping of research fields to uncover relationships, gaps, and emerging work, with a focus on structure and context rather than linear search results
& \begin{tabular}[t]{@{}l@{}}
Research Rabbit \\
Connected Papers \\
Litmaps
\end{tabular} \\

\bottomrule
\end{tabularx}
\end{table}
\subsection{Current AI tool usage and resulting problems}
The increasing availability of GenAI tools is reshaping research practices by embedding AI systems into everyday academic workflows. Rather than independently conducting all stages of analysis, synthesis and evaluation, researchers increasingly interact with AI systems to review and summarize literature, generate hypotheses, draft manuscripts, support data interpretation or refine arguments \cite{dwivediOpinionPaperWhat2023a}. AI-driven summarization, in particular, has been highlighted as a promising approach to support more efficient comprehension of scientific texts and informed decision-making \cite{shenekjiEvaluationDifferentAI2025}. Large language models (LLMs) such as \textit{GPT}-based systems, \textit{Claude}, or \textit{LaMDA} have demonstrated the ability to produce coherent, human-like text across a wide range of tasks \cite{thoppilanLaMDALanguageModels2022}. Beyond text generation, AI tools are also used to assist with programming and technical implementation, enabling users with limited prior expertise to produce functional code \cite{camperoTestEvaluatingPerformance2022}. 
In this context, GenAI tools for scientific research are increasingly framed as productivity-enhancing systems that may help researchers manage growing information volumes and research complexity. Tools like \textit{SciSpace} and \textit{Monica AI} were found to substantially reduce reading time while maintaining rapid comprehension of scientific literature \cite{shenekjiEvaluationDifferentAI2025}. By delegating repetitive and cognitively less demanding activities to AI systems, researchers could allocate greater attention to higher-level conceptual work.
However, despite the promises put forward by GenAI tool providers, critical questions remain regarding the reliability, limitations, and responsible use of these tools in scientific contexts. Systematic comparative evaluations of leading systems across core research tasks remain scarce, making it difficult to objectively assess progress and persistent gaps \cite{mikalefArtificialIntelligenceCapability2021}. At the same time, concerns about biases, inaccuracies, and misuse have largely remained conceptual. Empirical evidence on the types of errors these systems produce, the conditions under which they fail, and how outputs should be validated is still limited \cite{liMachineLearningConcrete2022}. As usage on GenAI-assisted workflows increases, the lack of benchmarks for the GenAI tools used is becoming increasingly problematic. Furthermore, existing assessments often focus primarily on technical performance (e.g. \cite{danlerQualityEffectivenessAI2024, featherstoneArtificialIntelligenceSearch2025, helmsandersenUsingArtificialIntelligence2025, laurentLABBenchMeasuringCapabilities2024}).
Although GenAI tools for scientific research are designed to support human researchers within complex academic workflows, many systems are not necessarily tailored to users’ cognitive and emotional needs. Beyond computational accuracy, factors such as comprehensibility of responses, usability, and adaptability to researchers’ working practices could be critical for effective and responsible adoption. Further, transparency is required to researchers informed about AI system activities and output generation. One key approach to address this challenge is explainable artificial intelligence (xAI). xAI refers to AI systems that are able to explain their rationale to human users, characterize their strengths and limitations and provide insight into their future behavior \cite{gunningDARPAsExplainableArtificial2019}. More specifically, xAI aims to make AI decision-making processes understandable by clarifying how, when, and why predictions or outputs are generated \cite{kamathExplainableArtificialIntelligence2021}. Consequently, xAI has the potential to strengthen user trust and support safer, more informed use of AI tools in scientific contexts \cite{kamathExplainableArtificialIntelligence2021}. This highlights the need not only for systematic benchmarking in general, but specifically for human-centered evaluation approaches that go beyond technical capabilities. Without such benchmarks, it remains unclear to what extent these systems truly fulfill their intended role in supporting scientific work. 
\subsection{Computer-centered benchmarks}
Existing computer-centered benchmarks primarily assess the technical performance of LLMs and GenAI tools across research tasks, focusing on accuracy, recall, reasoning and information retrieval. Laurent
\cite{laurentLABBenchMeasuringCapabilities2024} evaluated seven state-of-the-art LLMs across core biological research tasks. While LLMs showed relative strength in interpreting tables from scientific research papers, human experts outperformed them in most areas. Limitations were particularly evident in figure interpretation, reasoning over supplementary materials and interaction with domain-specific databases. Notably, models frequently refused to answer when required to retrieve specific information.
Complementing general task-based evaluations, Shenekji \cite{shenekjiEvaluationDifferentAI2025} provides a comparative assessment of GenAI tools across different functional categories relevant to scientific research. The findings indicate that even within the same functional categories, tools differ in how they implement and prioritize specific capabilities. In the domain of summarization, some GenAI tools for scientific research (e.g.: \textit{Quillbot, Jasper.ai}) emphasize efficient content condensation, often complemented by auxiliary features such as paraphrasing or grammar support, while others provide more basic summarization functionality with fewer enhancements. Similarly, among tools categorized as “chatting with PDF files” differences emerge in terms of interaction depth, and the extent to which they support detailed explanation and information extraction from scientific documents. These variations suggest that GenAI tools differ not only in performance, but also in how specific research support functions are operationalized. This in turn shapes their suitability for different research tasks.
While studies focus on document-centered interaction and Q\&A performance, other research shifts attention toward the use of GenAI tools for scientific research in structured literature review and evidence synthesis workflows. Assessments of widely used GenAI tools (e.g. \textit{Elicit, ChatGPT, Consensus, SciSpace}) indicate promising potential to facilitate review processes: They often generated high-quality responses, relied on scientific sources and remained consistent in citation \cite{danlerQualityEffectivenessAI2024}. Furthermore, GenAI tools including \textit{Elicit} and \textit{ChatGPT} correctly extracted around 90\% of required study data in systematic review settings, with particularly strong performance on basic study and participant characteristics \cite{helmsandersenUsingArtificialIntelligence2025}. These findings demonstrate considerable efficiency gains, including reductions in time and labor required for evidence synthesis. 
However, empirical evidence consistently reveals technical constraints. Hallucinated data, fluctuations in response quality, inconsistent retrieval behavior, and inclusion of non-scientific or irrelevant sources necessitate careful human verification \cite{danlerQualityEffectivenessAI2024, featherstoneArtificialIntelligenceSearch2025,  helmsandersenUsingArtificialIntelligence2025}. Complex searches remain challenging, with high variability in performance, even under expert prompting \cite{featherstoneArtificialIntelligenceSearch2025}. Consequently, effective use of these systems still depends on substantial methodological expertise and critical oversight.
Taken together, existing benchmarks indicate that GenAI systems demonstrate promising capabilities in structured tasks such as tabular interpretation, basic data extraction and literature summarization. At the same time, they reveal persistent weaknesses in complex reasoning, domain-specific database interaction, consistency and reliability. While these benchmarks provide valuable insights into technical capabilities, they offer limited understanding of how such systems function in real research workflows from a user perspective or how they are used and perceived. 
\subsection{Human-centered benchmarks}
Human-centered benchmarks extend beyond purely technical performance metrics and examine how GenAI tools are experienced and evaluated by users within real research workflows. In addition to output accuracy, these studies consider metrics like perceived usefulness, task relevance, usability, workflow integration, and suitability for specific scientific contexts.
Similar to computer-centered findings, human-centered evaluations indicate that current GenAI systems do not yet reach human-level performance across complex research processes. Morande \cite{morandeBenchmarkingGenerativeAI2023} argues that existing tools tend to specialize in isolated research tasks rather than providing comprehensive support across the full research life cycle. While they can effectively assist specific steps, no single, overarching AI solution currently meets the diverse and evolving demands of scientific research. Current GenAI tools often require multi-model workflows, where outputs from different systems must be manually combined. This fragmentation extends beyond the research process itself to the interfaces of GenAI tools, creating additional coordination demands for users \cite{buschekCollageNewWriting2024a}. At the same time, users’ roles are shifting from performing tasks directly to making editorial and compositional decisions about generated outputs. These changes call for user-interface and interaction designs that better support such emerging responsibilities \cite{buschekCollageNewWriting2024a}. To identify which tools and interface designs are actually suited to these changing requirements, their performance and usability must be evaluated in relation to the demands of specific research tasks.
For this reason, systematic benchmarking is needed to assess the suitability of particular GenAI tools for task-specific research contexts \cite{morandeBenchmarkingGenerativeAI2023}.
At the same time, other studies highlight a more optimistic perspective regarding multi-model workflows: Biju \cite{bijuDesigningAIDrivenSLR2025} decomposed the systematic literature review process into discrete stages and identified specialized AI tools for each step. Based on this mapping, they proposed a six-tool multi-model workflow, which was positively evaluated by student participants. A majority reported improved efficiency compared to traditional search engines such as Google, highlighting enhanced organization, validation mechanisms, reduced error rates and time savings. Tools such as \textit{Semantic Scholar}, \textit{Litmaps} and \textit{Scite.ai} were perceived as outperforming conventional search engines in supporting structured evidence synthesis. The study shows that a well‑linked set of tools can offset each other's weaknesses and speed up work, as long as they offer better search, higher accuracy, and are easy to use \cite{bijuDesigningAIDrivenSLR2025}.
Further human-centered comparative analyses assess not only output quality but also customization, integration, cost, response time, and domain suitability. Touati Hamad \cite{touatihamadComparativeAnalysisChatGPT2024} systematically evaluated and compared \textit{ChatGPT} and research GenAI tools across these criteria. While several systems demonstrated user-friendly interaction (e.g., \textit{ChatPDF}, \textit{AskYourPDF}), performance varied substantially in terms of customization and overall effectiveness. \textit{ChatGPT} emerged as particularly versatile and reliable, with strong integration capabilities and comparatively high accuracy. Nevertheless, even high-performing GenAI tools were described as requiring further improvement in contextual understanding, integration with research platforms, customization features and ethical safeguards.
Despite promising usability features, recurring concerns emerge regarding transparency, reproducibility and the reliability of generated outputs. Several studies report opaque source selection mechanisms in literature review tools, unclear ranking criteria for recommended literature and limited insight into how systems determine the “most important” research works \cite{chowdhuryAILLMdrivenSearch2024, danlerQualityEffectivenessAI2024, featherstoneArtificialIntelligenceSearch2025}. In many cases, GenAI tools produce incomplete or inaccurate citations, reference non–peer-reviewed materials such as company websites or blogs, or fail to provide verifiable bibliographic information \cite{chowdhuryAILLMdrivenSearch2024}. As a result, users are often unable to easily trace or validate the sources underlying generated responses. Limited reproducibility of results and insufficient transparency in retrieval and evidence synthesis processes reduce trustworthiness and necessitate manual verification, even when outputs appear useful or efficient \cite{chowdhuryAILLMdrivenSearch2024}.
\section{Research Gap}
Taken together, existing computer-centered and human-centered benchmarks provide valuable yet fragmented insights into the role of GenAI tools in scientific research. Computer-centered evaluations highlight strengths in structured sub tasks such as summarization and data extraction, but consistently reveal weaknesses in complex reasoning, systematic retrieval and reliability. Human-centered studies report perceived gains in efficiency and workflow support while simultaneously exposing transparency limitations, fragmented tool ecosystems and the continued need for methodological expertise and manual verification.
Despite these insights, current evaluations remain limited in several important ways. Although some studies compare GenAI tools for specific research tasks, systematic benchmarks in this domain remain relatively scarce, particularly from a human-centered perspective. The existing evidence base is also uneven across research functions: While literature review tools have received increasing attention, empirical evaluations of Q\&A tools remain comparatively limited. In addition, evaluation dimensions such as explainability, transparency, and reproducibility are frequently discussed but rarely operationalized as explicit benchmark metrics. Consequently, integrated evaluation approaches that combine technical performance indicators with human-centered criteria across realistic research tasks are still lacking.
At the same time, the number of available GenAI tools for scientific research and the number of researchers using them continues to grow rapidly \cite{dingRiseGenerativeArtificial2025}. As a result, a structured and user-focused evaluation framework becomes increasingly necessary to assess whether such systems are truly suitable for supporting core research activities.
To address this gap, this study proposes a human-centered benchmark approach that evaluates GenAI tools for scientific research across two common support functions: document-based question answering and AI-assisted literature review. Rather than focusing solely on model capabilities, the benchmarks assess tool performance across a set of task-oriented metrics reflecting typical researcher needs. By systematically comparing tools within these functional categories, the study aims to provide a structured assessment of their suitability for supporting research-related tasks. In addition, this research report describes qualitative observations regarding the usability of the GenAI tools.
Accordingly, this study addresses the following research questions:
\begin{researchq}
How can AI tools be evaluated from a human-centered perspective?
\end{researchq}
\begin{researchq}
Are AI tools suitable for supporting research-related tasks?
\end{researchq}
\section{Methods}
\subsection{Q\&A Tools}
We evaluated five  Q\&A tools: \textit{ScienceOS}, \textit{Humata}, \textit{ChatPDF}, \textit{PDF.ai} and \textit{AskYourPDF}. We chose these tools for several reasons:
\begin{itemize}
    \item all of them possess Q\&A functionality (answer questions based on provided documents)
    \item all of them have a clear focus on research papers (e.g. indicating research-focused use cases using testimonials)
    \item they incorporate AI to answer questions
    \item they were the first tools to appear when entering the keywords “chat”, “pdf” and “AI” on Google
\end{itemize}
We chose the fourth criterion, because data regarding real-world usage frequency was not available to us. We are aware however, that high search engine rankings are not only the product of popularity among users. All tools were used in the paid versions since they provided smaller limitations for paper uploads and allowed the free choice of LLMs.
\subsubsection{Materials}
To assess the capability of GenAI tools in answering questions about scientific documents, a corpus of research papers was selected. The evaluation was conducted using two different settings: a single-document chat, in which each tool was tested on one paper at a time, and a multi-document chat, in which multiple papers were provided simultaneously. The latter setting aimed to assess whether the tools were able to process and compare information across different documents.
For the multi-document chat, the selected papers were required to share key characteristics. Specifically, all documents were journal articles within the same subject area to ensure comparability. Based on the academic background of one of the evaluators, psychological studies on emotional processes in individuals with Borderline Personality Disorder were selected, as prior familiarity with the content facilitated a more reliable evaluation. The selected papers are listed in table \ref{tab:qa_tools_benchmarking_paper_corpus}.
\begin{table}[htbp]
\centering
\caption{Q\&A Tools Benchmarking: Paper Corpus}
\label{tab:qa_tools_benchmarking_paper_corpus}
\begin{tabularx}{\textwidth}{
    c
    >{\raggedright\arraybackslash}p{2.5cm}
    >{\raggedright\arraybackslash}p{3.2cm}
    >{\raggedright\arraybackslash}X
}
\toprule
\textbf{No.} & \textbf{Chat} & \textbf{Paper} & \textbf{Topic} \\
\midrule
1 & Single-document chat 
  & Haws \cite{hawsExaminingAssociationsPTSD2022}
  & Associations between PTSD symptoms and aspects of emotion dysregulation in traumatized women, using network analytic methodology \\
\addlinespace
2 & Multi-document chat 
  & Herpertz \cite{herpertzEmotionCriminalOffenders2001}
  & Emotional reactivity of criminal offenders with BPD and psychopathy to pleasant and unpleasant stimuli \\
\addlinespace
3 & Multi-document chat 
  & Matzke \cite{matzkeFacialReactionsEmotion2013}
  & Emotional recognition in persons with BPD \\
\addlinespace
4 & Multi-document chat 
  & Steinbrenner \cite{steinbrennerDecreasedFacialReactivity2022}
  & ``Social communication''-facial mirroring, emotional contagion, and emotion recognition-in persons with BPD \\
\addlinespace
5 & Multi-document chat 
  & Pizarro-Campagna \cite{pizarro-campagnaCognitiveReappraisalImpairs2023}
  & Ability to apply emotion regulation strategies, expressive suppression \& cognitive reappraisal, in a socially rejecting context in persons with early-stage BPD \\
\bottomrule
\end{tabularx}

\vspace{0.5em}
\begin{minipage}{\textwidth}
\small
\textit{Note.} PTSD = posttraumatic stress disorder; BPD = borderline personality disorder.
\end{minipage}
\end{table}
\subsubsection{Measures}
To evaluate the performance of Q\&A tools, a set of nine metrics was developed. The evaluation covered a range of document understanding capabilities (computer-centered metrics), including summarization \textit{(Single chat interpretation)}, figure interpretation \textit{(Image information description)} and the extraction of structured information such as tables, graphs or formulas \textit{(Extraction)}. In addition, the consistency between generated outputs and the original source documents was assessed \textit{(Consistency external)}. The evaluation also considered the range of document formats supported by each GenAI tool \textit{(Diverse document formats)}, as this directly affects their applicability in real research workflows.
Particular emphasis was placed on human-centered metrics, as these are central to assess the practical usefulness of the tools in research workflows. In this context, four key dimensions were examined: \textit{Intent match} refers to the extent to which the generated response addresses the underlying information need expressed in the prompt. To evaluate \textit{Intent match}, the answers of the GenAI tools to all six \textit{Consistency external} prompts were considered and aggregated into an overall score. \textit{No-answer criteria} capture whether the tool explicitly acknowledges missing information instead of generating unsupported or potentially hallucinated content. \textit{Multi-chat interpretation} evaluates the ability to identify similarities and differences across multiple documents in response to a given query. Finally, \textit{xAI accuracy} measures whether the GenAI tool correctly identifies and highlights the relevant text passages in the source documents that support its answer. \textit{xAI accuracy} was estimated retrospectively across the answers to all metrics.
To operationalize the evaluation, a set of standardized prompts was developed and applied consistently across all tools. For each metric, at least one prompt was designed and entered into each system under identical conditions, without allowing follow-up interactions. See table \ref{tab:prompt-examples} for examples. Depending on the evaluation setting, either a single document or multiple documents were provided as input. The generated responses were recorded and compared against predefined ground-truth answers. We defined these, based on the information contained in the papers after conducting a thorough analysis of the literature prior to the evaluation. 
\newcolumntype{Y}{>{\raggedright\arraybackslash}X}

\begin{table}[htbp]
\centering
\caption{Prompt examples}
\label{tab:prompt-examples}
\renewcommand{\arraystretch}{1.2}

\begin{tabularx}{\textwidth}{p{0.30\textwidth} Y}
\toprule
\textbf{Metric} & \textbf{Prompt example} \\
\addlinespace

Consistency External
& How can the phenomenon of adaptive emotion regulation be defined in a broader sense? \\
\addlinespace

No-answer criterion
& How does the Chatroom-Task work? \\
\addlinespace

Multi-chat Interpretation
& What are similarities and differences between the measures used in the papers to measure emotional responses to the stimuli? \\

\bottomrule
\end{tabularx}
\end{table}
All outputs were evaluated by a domain-informed rater using a five-point Likert scale ranging from 1 (incorrect or incomplete output) to 5 (correct and complete output). In cases where a tool did not provide an answer or produced irrelevant output, an additional no-answer category (X) was assigned.
In addition to the metric-based evaluation, qualitative observations regarding the usability of the tools were documented during the benchmarking process in order to capture practical aspects relevant to research workflows. Refer to table \ref{tab:q_and_a_metrics_all} in the appendix for a list of all incorporated metrics.
\subsubsection{Procedure}
After developing the metrics, prompts, and evaluation scheme, we selected the GenAI tools to be evaluated. To ensure better comparability, the GPT-4 model version was used across all GenAI tools. Two raters with a bachelor’s degree in psychology and sociology conducted the evaluation between June and August 2025. Each GenAI tool was evaluated by one of the two raters. The selected papers were uploaded and a new chat session was created in which all prompts were entered sequentially within the same conversation. All answers generated by the GenAI tools, as well as the raters’ observations, were documented in Excel spreadsheets. The evaluation of each GenAI tool was conducted over several days due to the limited weekly working hours of the raters.
After each prompt, the generated answer was compared with the predefined ground-truth answer. In this way, the GenAI tools were evaluated sequentially.
For the metrics \textit{Consistency external} and \textit{Multi chat interpretation}, overall scores were calculated by averaging the ratings across the corresponding prompts for each GenAI tool (\textit{Consistency external overall}, \textit{Multi chat interpretation overall}).
\subsection{Literature Review}
We evaluated four generalist AI tools with literature review capabilities: \textit{ChatGPT}, \textit{DeepSeek}, \textit{Perplexity AI} and \textit{You.com}. Further, we evaluated \textit{Consensus AI} as a specific literature review tool. We chose these tools for several reasons:
\begin{enumerate}
    \item all of them possess literature review capabilities (retrieving relevant literature based on a research question in natural language)
    \item they are capable of conducting web search to retrieve literature from relevant data bases
    \item they incorporate AI to retrieve literature and answer the research question
\end{enumerate}
\subsubsection{Materials}
To benchmark the literature review capabilities of the selected GenAI tools, a research question was defined and used consistently across all systems. The following question guided the literature search: “Does the process of Emotional Contagion differ between individuals with Borderline Personality Disorder (BPD) and healthy individuals?” This topic was selected because prior literature exploration on this question had been conducted by a member of the research team in 2023. Thus, proficiency in evaluating topic-related answers and defining the ground-truth was present. Using the same research question provided an initial foundation of previously identified literature and allowed the research team to extend and refine the existing search. The literature review was therefore repeated to identify more recent publications as well as potentially relevant papers that may not have been captured during the earlier search due to time constraints.
Research on Emotional Contagion is a relatively recent and evolving field. Consequently, several related terms were used to describe similar or overlapping constructs, including Affective Contagion, Emotional Transmission, Emotional Synchrony or Shared Emotions. This conceptual variability complicates systematic literature searches and therefore represents a suitable test case for evaluating AI-assisted literature review tools.
Based on the previous literature exploration and an updated manual search, a reference literature set consisting of 38 relevant publications published between 1999 and 2025 was created. This reference set was used to evaluate the completeness and relevance of the literature retrieved by the GenAI tools during the benchmarking process.
\subsubsection{Measures}
The benchmarking was conducted from three complementary perspectives: 
\begin{enumerate}
    \item supportive functionalities
    \item xAI features
    \item characteristics of the retrieved literature
\end{enumerate}
The selected measures were designed to capture both the technical retrieval performance of the GenAI tools and their usability within a research workflow. An overview of all measures used in the literature review benchmark is provided in table \ref{tab:litrev_metrics_all} in the appendix.
From a human-centered perspective, literature review tools should not only retrieve relevant publications but also support researchers throughout the search and analysis process. Therefore, the evaluation considered both the availability of supportive functionalities (functionalities that extend the core literature review functionality) and the presence of xAI features that improve transparency and interpretability. To define relevant evaluation criteria, consultations were conducted with the development team to identify functionalities considered useful in scientific research practice. Based on these discussions, the tools were examined for the presence of several supportive functionalities, which are listed in table \ref{tab:supportive-functionalities}. Any additional functionalities offered by the tools were documented. In addition, we assessed the presence of  xAI features. The measures selected for this purpose can be viewed in table \ref{tab:xai-features}.
\newcolumntype{Y}{>{\raggedright\arraybackslash}X}

\begin{table}[htbp]
\centering
\caption{Supportive functionalities and explanations}
\label{tab:supportive-functionalities}
\renewcommand{\arraystretch}{1.2}

\begin{tabularx}{\textwidth}{p{0.35\textwidth} Y}
\toprule
\textbf{Supportive functionality} & \textbf{Explanation} \\
\addlinespace
Web Search Selection
& Ability to manually toggle whether the tool is allowed to scrape the web for fitting sources. \\
\addlinespace
Knowledge database link
& Direct linking of retrieved sources to their original publications. \\
\addlinespace
Visualization of the retrieved literature
& Ability to show data in a visually appealing and easy-to-understand way, for example as a table or graph. \\
\addlinespace
Generation of a search query
& Ability to generate structured search queries for systematic literature reviews. \\
\addlinespace
Setting of search criteria and advanced search
& Ability to apply filter options to refine search results, as well as access advanced search modes, for example ``Research Mode'' or ``Deep Search''. \\
\addlinespace
Use of search operators in search options
& Ability to explicitly use Boolean search operators in a user's search prompt. \\
\addlinespace
Manual exclusion of literature
& Ability to manually exclude retrieved sources from the literature list the tool uses. \\
\addlinespace
Automated report generation
& Ability to generate an explicit report about the retrieved literature beyond the tool's user-interface output. \\

\bottomrule
\end{tabularx}
\end{table}

\newcolumntype{Y}{>{\raggedright\arraybackslash}X}

\begin{table}[htbp]
\centering
\caption{xAI features and explanations}
\label{tab:xai-features}
\renewcommand{\arraystretch}{1.2}

\begin{tabularx}{\textwidth}{p{0.35\textwidth} Y}
\toprule
\textbf{xAI feature} & \textbf{Explanation} \\
\midrule

Confidence scores
& Whether the tool's response was accompanied by a probabilistic score indicating how certain the given answer was to be correct. \\
\addlinespace
Explanation of score composition
& Whether there was an explanation for the given score, in cases where such a score was provided. \\
\addlinespace
Verifiability of results
& Whether the given output could be checked for truthfulness, which was possible for all tools by manually checking the provided sources and their quality. \\
\addlinespace
Transparency regarding searched databases
& Whether the tools made clear which databases or data sources were searched to find relevant literature. \\
\addlinespace
Progress indicator during search
& Whether there was any kind of visual progress indicator showing the human operator what the tool was currently doing, thereby increasing transparency regarding the search steps taken. \\
\addlinespace
Research report: exportability with explanations
& Whether the export function included relevant metadata for the given output, such as date of search, database, type of LLM, and related information. \\

\bottomrule
\end{tabularx}
\end{table}
The literature identified by each GenAI tool was evaluated using several measures focusing on both the quantity and quality of the retrieved sources. First, the total number of retrieved sources was recorded. Subsequently, the retrieved sources were analyzed with regard to the following characteristics: \textit{number of existing sources}, \textit{number of scientific sources}, \textit{number of peer-reviewed} sources and the degree of \textit{Intent match}, defined as the extent to which a source addresses the information need expressed in the research question. \textit{Intent match} was evaluated on a five-point Likert scale ranging from 1 (= source fails to fit the thematic context of user’s need) to 5 (= source fits the thematic context of user´s need). Sources with a score of 3 or higher were considered intent matching. Based on these criteria, the number of usable sources for answering the research question was determined. A source was defined as usable if it 
\begin{enumerate}
    \item existed (no hallucination by the LLM occurred)
    \item was scientific (source had to be published in a credible scientific journal or conference)
    \item was peer-reviewed (by said journal or conference)
    \item met the \textit{Intent match} threshold
\end{enumerate}
In addition, the \textit{Reproducibility} of the retrieved literature was assessed by conducting a second search trial in a new chat using the same setup and comparing the overlap between the retrieved source lists of trial one and two. \textit{Reproducibility} was quantified using the Jaccard-Index (score ranging from 0\% to 100\%). This measure reflects the intersection of the results that are the same across both chats in relation to all results obtained.
\subsubsection{Procedure}
After developing the research question, reference literature, prompts, and evaluation scheme, we selected the GenAI tools to be evaluated. In \textit{ChatGPT}, GPT-5, was used with “Deep Research” and “Web Search” turned on. In \textit{DeepSeek} the Model V3.2 was used with “Deep Thinking” and “Web Search” activated. \textit{Consensus AI} was used in “Deep Search” mode and for \textit{You.com} “Research Mode” and “Deep Search” were used. Otherwise both \textit{Consensus AI} and \textit{You.com} were tested with default settings. One rater with a bachelor’s degree in sociology conducted the evaluation between November 2025 and January 2026. Each GenAI tool was evaluated by the rater. The research question was entered twice per tool each in a new conversation to calculate \textit{Reproducibility}. For all metrics except \textit{Reproducibility}, the results of the first chat were used for evaluation. After the first prompt, no further prompts were entered. All of the rater’s observations and evaluations were saved in Excel spreadsheets. The evaluation of each GenAI tool was conducted over several days due to the limited weekly working hours of the rater. After retrieving the literature, it was exported using one of the formats offered by each tool, e.g. csv.
\section{Results}
\subsection{Q\&A tools}
Across all metrics, performance varied substantially between tools, but also within individual tools across the single and multi document chat settings. An overview of the results is provided in tables \ref{tab:q_and_an_computer_centered} and \ref{tab:q_and_a_human_centered}. 
\subsubsection{Computer-centered metrics}
The evaluated GenAI tools showed their strongest performance in \textit{Image information description} in the single-document chat setting, where most tools achieved a score of 5, with the exception of \textit{ScienceOS}. Interestingly, \textit{ScienceOS} performed best in \textit{Image information description} in the multi-document setting, while several other tools produced incorrect or incomplete responses or failed to provide an answer. With regard to \textit{Consistency external}, most GenAI tools demonstrated relatively strong performance, with an overall score of 4 when \textit{AskYourPDF} (scores of 2 and 3) was excluded. \textit{Single chat interpretation} also showed generally good performance, although scores varied considerably between GenAI tools, ranging from 2 to 5.
In contrast, the evaluation of \textit{Extraction} revealed particularly inconsistent results. While \textit{Extraction of formulas} performed well in the multi-document chat setting (only \textit{ScienceOS} and \textit{PDF.ai} failed to produce responses), four out of five GenAI tools did not provide an answer in the single-document setting. \textit{Extraction of table information} showed substantial variation across GenAI tools: Performance ranged from missing or incomplete answers to highly accurate responses, with noticeable differences both between tools and between the single- and multi-document settings. Across all metrics, the weakest performance was observed for \textit{Extraction of graph information}. In the multi-document setting, none of the evaluated GenAI tools produced a response for this task.
Considerable differences were also observed regarding supported document formats. While some tools only accepted PDF files, others supported a broader range of formats including DOCX, PPTX, XLSX, CSV or image files. \textit{ChatPDF} demonstrated the highest level of format diversity, whereas \textit{ScienceOS} and \textit{PDF.ai} were limited to PDF documents. See table \ref{tab:q_and_an_computer_centered} for an overview.
\newcolumntype{Y}{>{\raggedright\arraybackslash}X}

\begin{table}[htbp]
\centering
\caption{Selection of Computer-centered Q\&A metrics}
\label{tab:q_and_an_computer_centered}
\renewcommand{\arraystretch}{1.2}

\begin{tabularx}{\textwidth}{*{6}{Y}}
\toprule
\textbf{Metric} & \textbf{Humata} & \textbf{AskYourPDF} & \textbf{scienceOS} & \textbf{PDF.ai} & \textbf{ChatPDF.com}\\
\midrule
Diverse document formats
& PDF, DOCX, XLSX & PDF, PPTX,
DOCX, CSV & PDF & PDF & PDF, DOCX,
PPTX, XLSX,
CSV, JPG, PNG
 \\
\addlinespace
Document summary* & 5 & 4 & 5 & 5 & PDF, DOCX, 
PPTX: 5
XLSX, CSV, 
JPG, PNG: X \\
\bottomrule
\end{tabularx}
\vspace{0.5em}
\begin{minipage}{\textwidth}
\small
\textit{Note.} *Prompt: What is the document about?; 1 = incorrect or incomplete output; 5 = correct or complete output; X = tool does not give any answer.
\end{minipage}
\end{table}
\subsubsection{Human-centered metrics}
The evaluation of \textit{Intent match} showed moderate and relatively similar performance across most GenAI tools. Overall scores ranged from 3 to 4, indicating that responses to \textit{Consistency external} prompts fitted the thematic context of user´s need partially. For example, when asked about previous scientific findings, the answers of most GenAI tools generally aligned with the prompt’s topic, but specific studies described in the paper were either omitted entirely or replaced with other studies mentioned in the paper that were unrelated to the prompt´s topic. \textit{PDF.ai} achieved the highest score (4), suggesting slightly better alignment with the information needs expressed in the prompts. All other tools received a score of 3, indicating responses that partially fitted the thematic context of user´s information need. 
\par
A clear difference emerged regarding the no answer criterion. While \textit{PDF.ai} and \textit{ChatPDF} frequently acknowledged when relevant information could not be retrieved from the uploaded document (scores ranged from 4 to 5), most of the tools (\textit{Humata}, \textit{AskYourPDF}, \textit{ScienceOS}) rarely did so and instead tended to generate answers based on external sources without referencing them (score of 1).
\par
Performance in \textit{Multi chat interpretation} was generally limited. Scores ranged between 1 and 3, indicating difficulties in identifying and comparing similarities and differences across multiple papers. Accordingly, comparison was often incomplete or imprecise: Key details were sometimes omitted, while other points were irrelevant or presented ambiguously, such as comparisons that did not clearly indicate which papers were referenced. Furthermore, answers of GenAI tools occasionally mixed correct and incorrect information. \textit{ScienceOS} and \textit{ChatPDF} achieved the highest scores (3), whereas \textit{PDF.ai} showed the lowest performance (score of 1). The responses generated by \textit{PDF.ai} did not correspond to the user’s intent, as no direct comparison between the papers was provided. Furthermore, it remained unclear whether all papers were included in the comparison and, in some cases, from which paper the information was retrieved. In addition, \textit{Humata} appeared unable to correctly process one of the papers, as it was never cited or included in the comparison.
\par
With regard to explainability, the evaluated tools showed generally limited performance. \textit{xAI accuracy} scores ranged between 1 and 3, indicating that highlighted passages often did not reliably correspond to the information used to generate the responses. \textit{ScienceOS} and \textit{ChatPDF} achieved the highest scores (3), while \textit{Humata}, \textit{AskYourPDF} and \textit{PDF.ai} showed weaker performance.
A closer inspection of the highlighted passages revealed several recurring issues. In many cases, markings were inconsistent with the generated answers, appeared random or were partially identical across different prompts. Some tools highlighted entire pages or irrelevant sections such as reference lists instead of specific supporting passages.
Tool-specific patterns were also observed. For example, \textit{ScienceOS} occasionally produced markings on pages different from those referenced in the response. \textit{ChatPDF} showed highly variable performance, with both accurate paragraph-level highlights and irrelevant markings. \textit{Humata} often highlighted only the final parts of sentences, sometimes with slightly shifted or incomplete markings. \textit{AskYourPDF} occasionally failed to link to highlighted passages correctly and when links were available, entire pages were sometimes indicated. \textit{PDF.ai} showed the weakest explainability performance, frequently producing excessive and seemingly random highlights, even when no answer was provided.
\begin{table}[htbp]
\centering
\caption{Human-centered Q\&A metrics}
\label{tab:q_and_a_human_centered}
\renewcommand{\arraystretch}{1.2}

\begin{tabularx}{\textwidth}{>{\raggedright\arraybackslash}X *{5}{>{\centering\arraybackslash}X}}
\toprule
\textbf{Metric} & \textbf{Humata} & \textbf{AskYourPDF} & \textbf{scienceOS} & \textbf{PDF.ai} & \textbf{ChatPDF.com} \\
\midrule
Intent match & 3 & 3 & 3 & 4 & 3 \\
\addlinespace
No answer criterion: single-document & 1 & 1 & 1 & 5 & 5 \\
\addlinespace
No answer criterion: multi-document & 1 & 1 & 1 & 5 & 4 \\
\addlinespace
Multi-chat interpretation & 2 & 2 & 3 & 1 & 3 \\
\addlinespace
xAI accuracy & 2 & 1 & 3 & 1 & 3 \\
\bottomrule
\end{tabularx}
\vspace{0.5em}
\begin{minipage}{\textwidth}
\small
\textit{Note.} 1 = response fails to fit the thematic context of user´s need / tool hallucinates / no differences or similarities are recognized / whole section or page is marked or relevant information is not included in the marked text; 5 = response fits the thematic context of user´s need / tool clearly states that it could not find the required information in the document / all differences or similarities are recognized / the whole marked paragraph is relevant for the given answer and relevant information is included in the marked text; X = nothing is marked  / tool does not give any answer / nothing is marked.
\end{minipage}
\end{table}
\par
In addition to the metric-based evaluation, several usability-related aspects became apparent during the interaction with the evaluated tools. These observations concern practical features that may influence how researchers organize documents, navigate conversations and trace the sources of generated responses.
\par
The evaluated tools differed in how uploaded documents could be organized within the interface. \textit{Humata}, \textit{ChatPDF} and \textit{AskYourPDF} offered folder structures that allowed documents to be grouped; however, on \textit{AskYourPDF}, uploaded documents were displayed across multiple folders at the same time, which limited the effectiveness of this feature. In contrast, \textit{ScienceOS} organised papers using “collections”, while \textit{PDF.ai} relied on a tag-based system. Issues were observed in the upload process: In \textit{PDF.ai}, papers could only be uploaded individually, as attempts to upload multiple documents sometimes resulted in an error message. In addition, \textit{PDF.ai} showed comparatively long upload times for documents requiring OCR processing.
\par
Regarding navigation and interaction with generated responses, access to previous conversations also varied between tools. In \textit{Humata} and \textit{PDF.ai}, earlier chats remained accessible when reopening the corresponding paper from the document library. \textit{ScienceOS} additionally allowed quick access to previous interactions via the sidebar interface. How to create a new chat in \textit{PDF.ai} remained unclear. Interaction with generated responses differed as well: In all GenAI tools except for \textit{PDF.ai}, responses could be copied easily with a single click. \textit{ChatPDF} provided comparatively concise responses by default, with the option to expand the generated output.
\par
Further differences concerned how sources supporting the generated answers were presented. In \textit{Humata} and \textit{ScienceOS}, links to specific pages were often provided directly after text segments, with sources listed again below the response. \textit{PDF.ai}, for example, displayed references only as page numbers below the response, which occasionally made it difficult to identify the corresponding document. Across several tools, instances of incomplete or inaccurate source attribution were observed.
\subsection{Literature review}
Literature review metrics were categorized into three domains: “supportive functionalities”, “xAI features” and “characteristics of retrieved literature”. The following outlines the tested tools results for each section, comparing noteworthy findings of section items. 
\par
The supportive functions category assessed basic functions of the tested tools besides their literature search functionality.   
Regarding \textit{Web search selection} all tested tools allowed the user to manually toggle web search for additional and more up-to-date information extraction, except for \textit{Consensus AI}, which only employed semantic scholar. All tools linked the source they were referring to in an easily accessible way (\textit{Knowledge database link}). Usually a sidebar view was used to show all the source links in list format. 
Every tool was at least able to generate a table representation of it’s literature findings, although this visual representation \textit{(table, graph, etc.)} wasn’t always chosen. There were runs in which the output was purely textual. None of our tested tools offered dedicated functionality to generate a search query. Only the tool DeepSeek didn’t offer additional functionality to \textit{set search criteria and advanced search}, besides its reasoning feature “deep thinking”. It was not possible  to manually use boolean operators in the program settings for any of the tools tested. Manual filtering of the found sources was not possible for four out of five tools. Only \textit{You.com} had this setting available. \textit{Report generation} was provided by \textit{Consensus AI} and \textit{Perplexity AI}. Otherwise, no separate report generation feature was observed. Table \ref{tab:litrev_supp_functions} shows the measurement items of this category and their binary results.
\begin{table}[htbp]
\centering
\small
\caption{Literature review supportive functions}
\label{tab:litrev_supp_functions}

\renewcommand{\arraystretch}{1.2}
\setlength{\tabcolsep}{4pt}

\begin{tabularx}{\textwidth}{@{}
  >{\raggedright\arraybackslash}X
  >{\centering\arraybackslash}p{1.3cm}
  >{\centering\arraybackslash}p{1.3cm}
  >{\centering\arraybackslash}p{1.5cm}
  >{\centering\arraybackslash}p{1.3cm}
  >{\centering\arraybackslash}p{1.5cm}
@{}}
\toprule
\textbf{Anchors} &
\textbf{You.com} &
\textbf{ChatGPT} &
\textbf{Consensus AI} &
\textbf{DeepSeek} &
\textbf{Perplexity AI} \\
\midrule

Web Search Selection &
Yes & Yes & No & Yes & Yes \\
\addlinespace
Knowledge Database Link &
Yes & Yes & Yes & Yes & Yes \\
\addlinespace
Visualization of the literature found (table, graph, etc.) &
Yes & Yes & Yes & Yes & Yes \\
\addlinespace
Generation of a search query &
No & No & No & No & No \\
\addlinespace
Setting of search criteria and advanced search &
Yes & Yes & Yes & No & Yes \\
\addlinespace
Use of search operators in search options &
No & No & No & No & No \\
\addlinespace
Manual exclusion of literature (source filter) &
Yes & No & No & No & No \\
\addlinespace
Report generation &
No & No & Yes & No & Yes \\
\bottomrule
\end{tabularx}
\vspace{0.5em}
\begin{minipage}{\textwidth}
\small
\textit{Note.} All evaluations are binary (yes/no) and tested whether a tool provided a built-in feature that matched the anchor description.
\end{minipage}
\end{table}
\par

xAI features examined the availability of explainable AI (xAI) features the tested tools had to offer.  As of now, no tool could produce \textit{Confidence scores} and therefore score explanations were not available as well.  Contrary to confidence scores, every tool provided a link to each of their sources found and used. This allowed for verification of results by manually checking the provided sources and their quality. Regarding \textit{Transparency of searched databases}, verification of databases that were used for the search was not possible. None of the tools displayed any information in this regard in their user interfaces. At the time of our test, most of the tools displayed a visual search progress indication (e.g. progress bar) in their interface though. Out of our five tested tools, \textit{DeepSeek} was the only one without this feature built-in.  For the last metric in this domain, \textit{Research report: exportability with explanations (date of search, database, type of LLM, etc.)}, it can be said that every LLM allowed to download a copy or summary of the answer given by the AI. The documents exported from the tool in this way did not contain any information about the date of the search, the databases searched, keywords used for the search,  type and version of the large language model used to generate the answer, nor which account executed the search.
Table \ref{tab:litrev_xai_functions} shows the measurement items of this category and their binary results.
\begin{table}[htbp]
\caption{Literature review xAI functions}
\label{tab:litrev_xai_functions}
\centering
\small
\begin{tabularx}{\textwidth}{@{}
  >{\raggedright\arraybackslash}X
  >{\centering\arraybackslash}p{1.3cm}
  >{\centering\arraybackslash}p{1.3cm}
  >{\centering\arraybackslash}p{1.5cm}
  >{\centering\arraybackslash}p{1.3cm}
  >{\centering\arraybackslash}p{1.5cm}
@{}}
\toprule
\textbf{Anchors} & \textbf{You.com} & \textbf{ChatGPT} & \textbf{Consensus AI} & \textbf{DeepSeek} & \textbf{Perplexity AI} \\

Confidence Scores & No & No & No & No & No \\
\addlinespace
Explanation of Score Composition & No & No & No & No & No \\
\addlinespace
Verifiability of Results & Yes & Yes & Yes & Yes & Yes \\
\addlinespace
Transparency regarding searched databases & No & No & No & No & No \\
\addlinespace
Progress indicator during search (transparency regarding search steps taken) & Yes & Yes & Yes & No & Yes \\
\addlinespace
Research Report: exportability with explanations (date of search, type of LLM, etc.) & No & No & No & No & No \\
\bottomrule
\end{tabularx}
\vspace{0.5em}
\begin{minipage}{\textwidth}
\small
\textit{Note.} All evaluations are binary (yes/no) and tested whether a tool provided a built-in feature that matched the anchor description.
\end{minipage}
\end{table}
\par
Besides tool functionalities we examined the retrieved sources. The analyzed data unveiled that three out of the five tested tools listed sources multiple times. These source duplicates usually had a different Uniform Resource Locator (URL) but their content was identical. \textit{You.com} found 82 sources in the first run, of which only 23 remained after removing all duplicate entries. \textit{ChatGPT} surfaced 17 sources of which 9 remained after screening. \textit{Perplexity AI} produced 71 sources and 68 remained. \textit{DeepSeek} did in fact not produce any sources when not directly prompted to do so. The standardized prompt used for this analysis did not include explicit instructions to provide sources. Therefore, \textit{DeepSeek} had to be excluded from the analysis. \textit{Consensus AI} was the only tool that produced a list of duplicate free entries. The exact source numbers for each tool can be found in the appendix as well. Furthermore, the data shows a huge gap between the sources that were found by the tools to generate their outputs and sources that were classified as “usable” by the human reviewer. Sources that matched all criteria were considered usable sources. For \textit{You.com}, of the 23 duplicate-free sources, 19 were considered usable. \textit{ChatGPT} ended up with seven usable sources, \textit{Perplexity AI} with only 15 of it’s duplicate-free 68 sources and \textit{Consensus AI} with only 14 of it’s 50 duplicate-free sources. It should further be noted, that some of the sources listed by the tools were published in predatory journals. This could put their scientific quality further into question. Please refer to table \ref{tab:litrev_litrev_measures} for an overview of retrieved sources and filtered sources.
In order to assess the reliability of a tool,  a second run using the same setup was conducted. This was done to control if sources that were assumed to be usable in the first run, were also found again and reclassified as usable in the second run. The Jaccard-Index was used to calculate the number of overlapping sources found between trial one and trial two. The calculation was performed with lists of found sources after duplicate screening. Using this method, \textit{You.com} achieved a Jaccard Index Score of 17.6\%, \textit{ChatGPT} of 25.0\%, \textit{Consensus AI} scored the best with 28.0\% and \textit{Perplexity AI} the worst with 11.8\% overlap. Since \textit{DeepSeek} did not give any sources, the Jaccard Index could not be calculated.
\begin{table}[htbp]
    \centering
    \caption{Literature measures}
    \label{tab:litrev_litrev_measures}
    \begin{tabularx}{\textwidth}{%
        l                                   
        >{\centering\arraybackslash}X      
        >{\centering\arraybackslash}X      
        >{\centering\arraybackslash}X      
        >{\centering\arraybackslash}X      
        >{\centering\arraybackslash}X}     
        \toprule
        \textbf{Tool name} &
        \multicolumn{5}{c}{\textbf{Number of sources found}} \\
        \cmidrule(lr){2-6}
        & \textbf{retrieved} & \textbf{original} & \textbf{scientific} &
          \textbf{intent‑matching} & \textbf{usable} \\
        \midrule
        you.com          & 82 & 23 & 19 & 19 & 19 \\
        \addlinespace
        ChatGPT          & 17 &  9 &  9 &  7 &  7 \\
        \addlinespace
        DeepSeek         &  0 &  0 &  0 &  0 &  0 \\
        \addlinespace
        Perplexity AI    & 71 & 68 & 63 & 15 & 15 \\
        \addlinespace
        Consensus AI     & 50 & 50 & 50 & 14 & 14 \\
        \bottomrule
    \end{tabularx}
    \label{tab:tool-sources}
    \begin{minipage}{\textwidth}
    \small
\textit{Note.} The number of retrieved and filtered sources for each tool.
    \end{minipage}
\end{table}
\section{Discussion}
\subsection{Summary \& placement in the literature}
In this study we used a benchmarking scheme with the focus on human-centered metrics to assess AI-based Q\&A and literature review tools. In addition, we added several computer-centered metrics, which were focused on the provided functions and options. By using this approach, we extend existing schemes by assessing the question whether the system output is usable for researchers. Specifically metrics such as \textit{Intent match} are centered around what the user expects from the output in addition to its accuracy. One caveat of this approach is that human-centered metrics have a higher degree of subjectivity in their ratings. However, when evaluating tools that should be used by humans, such metrics cannot be neglected.
Overall, the results of our benchmark draw a mixed picture of GenAI tool´s suitability for research. Regarding Q\&A tools, a metric that showed comparatively high scores was \textit{Consistency external}, which describes the degree of consistency of the output with the original research paper. Further, tools showed good performance in creating holistic and correct summaries \textit{(Single-chat interpretation)}. On the other hand, tools struggled with \textit{No answer criteria} (meaning they did not admit on not finding certain information). This is a current phenomenon related to hallucinations in LLMs \cite{alansariLargeLanguageModels2026}. With regard to computer-centered metrics, such as information \textit{Extraction}, there was a great variability. This concerned for instance the extraction of formulas, graph points or image information. This is consistent with research stating that error rates when extracting data can reach a high level without human supervision \cite{mikriukovAIToolsAutomating2025}. Interestingly, this variability was not only present between tools but also between single-document chat and multi-document chat. Lastly, although the Q\&A tools produced reasonably accurate outputs, their \textit{xAI accuracy} was low. This means, that the highlighted passages in the source document were often irrelevant, or relevant passages were missed entirely. Indeed the faithfulness of xAI explanations varies, e.g. depending on the used method \cite{dehdariradEvaluatingExplainabilityLanguage2025}. This creates a critical verification problem: if the highlighted evidence does not reliably correspond to the generated answer, researchers cannot efficiently validate whether an output is correct. Without trustworthy explainability, the burden of verification falls back on the researcher, potentially making it faster to skim the source paper manually using a search function than to cross-check the tool's reasoning. We therefore recommend using Q\&A tools only for gaining a quick overview of a paper, rather than for extracting specific pieces of information where accuracy must be confirmed. An additional advantage of Q\&A tools might be the adaptability of the output, which can be achieved by natural language. For instance, wording and structure of the tool’s output can be manipulated.
Regarding literature review tools, a small amount of xAI features were present. All of the tools made the sources verifiable, for example by providing hyperlinks. Furthermore, during the review process, most tools provided a search state progress indicator that displayed the tool’s current activities. However, there was no transparency about the searched databases or reasoning for choosing specific databases or papers. Mikriukov \cite{mikriukovAIToolsAutomating2025} further argue that most commercial tools use close-source LLMs, making the processes even more opaque. Most importantly, even though the number of searched and cited sources was high in some tools (e.g. \textit{You.com}), the amount of sources that were usable (peer-reviewed, scientific source and matching user intent) was significantly lower. Also, \textit{Reproducibility} across repeated runs with identical prompts was low. This makes literature review tools unusable for systematic reviews. Other research confirms that the number of retrieved sources can vary greatly between different runs \cite{cassellAnalysisArticleScreening}. Such tools could be used however for unsystematic search processes or in addition to deterministic approaches (searching databases manually using boolean search strings). Opposed to databases, AI-based literature review tools provide the advantage of prompting the system in natural language instead of boolean search strings. Thus, they could be used to discover literature which was not apparent in manual search processes.
Taken together, both benchmarked tool types reveal a similar pattern: while GenAI tools can improve efficiency in early stages of the research workflow, their outputs still require careful human verification due to limitations in explainability, reproducibility, and source transparency.
\subsection{Limitations}
Although our benchmarking provides valuable insights into AI-supported Q\&A and literature review tools, there are several limitations to our approach.
Firstly, due to constraints in resources, each tool was only evaluated by one rater. This is a significant source of bias, especially for subjective metrics, such as usability measures. Further, some metrics are crucial, for the human-centered assessment of AI tools but at the same time they hold greater potential for subjectivity than others.  For instance it is important that the users’ “intent” is  met by the AI-system. At the same time, the intent or informational need behind the same prompt may vary greatly from user to user. However, several objective metrics, such as duplicate count are less prone for bias and human error.
Secondly, the AI-tool landscape is rapidly evolving. While our results might have been up to date during assessment, tool performance might have enhanced in the meantime. Thus, there is a need for reassessing evaluated tools and investigating new tools. Especially LLMs which answer questions in Q\&A tools, or find literature matching a natural language prompt are constantly evolving.
Thirdly, the “ground truth” to measure \textit{Consistency external} was complex and multifaceted. This means, that the determination if the output was true or false incorporated some room for interpretation.  For that reason, we chose an ordinal Likert scale, rather than a binary yes or no measure. Additionally, LLM outputs need to be evaluated holistically and semantically because they communicate in dialogues.
Lastly, the prompt in the literature review benchmarking was short and due to comparability reasons we did not allow for refinements in further prompts. In the real world however, users would probably specify their prompt, if they did not receive the desired output. One approach to overcome this issue in future benchmarking could be to provide a single-but more detailed prompt, e.g. stating that scientific resources should be included opposed to solely providing the research question. On the other hand, the user intent to receive research papers should be recognized by AI tools, when confronted with a research question.
\subsection{Future research}
Future research should apply our approach to the same types of research tools involving multiple raters. Further, results are highly dependent on LLMs which are constantly evolving regarding their performance. Thus, benchmarking tools with current models could yield different results, even though problems such as the lack of transparency, hallucinations and little reproducibility might still be present due to the probabilistic nature of LLMs. Further, for the literature review tools, it might be valuable to check retrieved sources not only for intent match and peer review but further quality criteria like the absence of predatory journals or assessment of journal metrics. Lastly, the additional integration of more computer-centric metrics into the framework might draw a more holistic picture of the tools.
\section{Conclusion}
In this study, we developed and applied a benchmarking framework combining human-centered and computer-centered metrics to evaluate AI-based Q\&A and literature review tools for research use. The results show that Q\&A tools can provide useful overviews and generally accurate summaries, but remain unreliable for precise information extraction due to hallucinations and weak explainability. Literature review tools were helpful for exploratory searches, yet their low reproducibility, limited transparency, and inconsistent source quality make them unsuitable for systematic reviews. Our main contribution lies in the human-centered focus of of the evaluation, which assesses whether AI tool outputs are actually usable for researchers in practice rather than evaluating technical performance alone.
\printbibliography
\newpage
\appendix
\section{Q\&A Measures}
\begin{table}[htbp]
\caption{Benchmarking of Q\&A tools: Investigated metrics}
\label{tab:q_and_a_metrics_all}
\centering
\begin{tabularx}
{\textwidth}{>{\raggedright\arraybackslash}p{3.2cm}
                                >{\raggedright\arraybackslash}X
                                >{\raggedright\arraybackslash}X}
\toprule
\textbf{Metric} & \textbf{Description} & \textbf{Evaluation}\\
\midrule

Consistency external &
Is the text information of a paper consistent with the produced output? &
5 = word-by-word consistency / correct summary \newline
4 = \newline
3 = no word-by-word consistency / meaning remains the same \newline
2 = \newline
1 = differences between paper and output in terms of meaning / hallucinations \\

\addlinespace

Intent match &
Does the answer state things within the thematic scope of user’s information need? \newline
(for consistency external only) &
5 = response fits the thematic context of user’s need / the prompt \newline
4 = \newline
3 = response fits the thematic context of user’s need / the prompt partially \newline
2 = \newline
1 = response fails to fit the thematic context of user’s need / the prompt \\

\addlinespace

No-answer criteria &
Does the answer state clearly if the required information was not found within the provided document and refrains the tool from using other sources? &
5 = tool clearly states that it could not find the required information in the document \newline
4 = \newline
3 = tool provides confusing feedback, e.g. tool claims that required information is in the document, but the source cannot be found \newline
2 = \newline
1 = tool hallucinates, i.e. lies about the information source \newline
X = tool does not give any answer \\

\addlinespace

Single chat interpretation &
How well can the paper be summarised? &
5 = summary contains all important information / is completely correct \newline
4 = \newline
3 = summary omits important information / is only partially correct \newline
2 = \newline
1 = summary is completely incorrect / contains only hallucinations \newline
X = unrelated output \\
\addlinespace

Multi chat interpretation &
How well can similarities and differences between the papers be identified? &
5 = all differences are recognized / all similarities are recognized \newline
4 = \newline
3 = some differences are recognized / some similarities are recognized \newline
2 = \newline
1 = no differences are recognized / no similarities are recognized \newline
X = unrelated output \\

\bottomrule
\end{tabularx}
\end{table}

\begin{table}[htbp]
\centering
\begin{tabularx}
{\textwidth}{>{\raggedright\arraybackslash}p{3.2cm}
                                >{\raggedright\arraybackslash}X
                                >{\raggedright\arraybackslash}X}
\midrule
\textbf{Metric} & \textbf{Description} & \textbf{Evaluation} \\
\midrule

Extraction \newline
(tables, formulas, graphs) &
How accurately can table data be extracted? &
5 = extracted data is correct / complete / same as in the paper \newline
4 = \newline
3 = extracted data is only partially correct / extraction of only certain values is possible \newline
2 = \newline
1 = extracted data is incorrect / incomplete / not from the paper \newline
X = no table information extraction possible / tool does not give any answer \\

& Can any formulas be extracted? &
5 = extracted formula is correct / the same as in the paper \newline
4 = \newline
3 = extracted formula is basically the one from the paper, but not complete / only partially correct \newline
2 = \newline
1 = extracted formula is completely incorrect / not the same as in the paper \newline
X = no formula extraction possible / tool does not give any answer \\

& How accurately can numerical values be extracted from graphs? &
5 = numerical extraction is completely correct / identical to the paper graph \newline
4 = \newline
3 = numerical extraction is partly correct, minor deviations from the paper graph \newline
2 = \newline
1 = numerical extraction is not correct / deviates from the paper graph \newline
X = no numerical extraction is possible / tool does not give any answer \\

\addlinespace

Image information description &
How exact can the image be described? &
5 = description of the image is correct \newline
4 = \newline
3 = description of the image is only partially correct \newline
2 = \newline
1 = description of the image is incorrect \newline
X = no image information extraction possible / tool does not give any answer \\

\addlinespace

Diverse document formats &
How well is the model's capability to handle various document formats? &
List formats that can be handled by the model, e.g. PDF, XLSX, DOCX, PPTX, CSV, JPG, PNG \newline
X = model-specific file format limitation / file format is not applicable \\

\addlinespace

XAI accuracy &
How easily can the paper paragraph be identified from which the answer was derived? &
5 = the whole marked paragraph is relevant for the given answer and relevant information is included in the marked text \newline
4 = \newline
3 = highlights more sentences than needed, but you can find the answer to the question within the marked text / relevant information is only partly included in the marked text \newline
2 = \newline
1 = whole section or page is marked / relevant information is not included in the marked text \newline
X = nothing is marked \\

\bottomrule
\end{tabularx}
\end{table}
\newpage
\section{Literature Review Measures}
\begin{table}[htbp]
\caption{Benchmarking of literature review tools: investigated metrics}
\label{tab:litrev_metrics_all}
\centering
\small
\renewcommand{\arraystretch}{1.2}

\begin{tabularx}{\textwidth}{%
  >{\raggedright\arraybackslash}p{2.6cm}
  >{\raggedright\arraybackslash}p{3.0cm}
  >{\raggedright\arraybackslash}X
  >{\raggedright\arraybackslash}X
}
\midrule
\textbf{Category} & \textbf{Metric} & \textbf{Description} & \textbf{Evaluation} \\
\midrule

Basic functions
& Supportive functions
& Are there functions beyond literature search?
& Tick if the following functions are available:
\begin{itemize}
    \item web search selection
    \item knowledge database link
    \item visualisation of the literature found, e.g. table, graph, etc.
    \item generation of a search query and, if possible, suggestions for improving search queries
    \item listing of excluded literature, i.e. false positives, possibly with an explanation of why literature was included or excluded
    \item setting of search criteria and advanced search
    \item use of search operators
    \item manual exclusion of literature, i.e. source filter
    \item report generation
\end{itemize}
List additional functions.

X = no additional functions
\\
\addlinespace
& XAI features
& Which XAI features does the tool have?
& Tick if the following features are available:
\begin{itemize}
    \item confidence scores and explanation of score composition
    \item verifiability of results
    \item transparency regarding searched databases
    \item progress indicator during search, i.e. transparency regarding search steps taken
    \item research report exportability with explanations, e.g. date of search, database, type of LLM, etc.
\end{itemize}
List additional XAI features.

X = no XAI features provided
\\
\bottomrule
\end{tabularx}
\end{table}

\begin{table}[htbp]
\centering
\small
\renewcommand{\arraystretch}{1.2}

\begin{tabularx}{\textwidth}{%
  >{\raggedright\arraybackslash}p{2.6cm}
  >{\raggedright\arraybackslash}p{3.0cm}
  >{\raggedright\arraybackslash}X
  >{\raggedright\arraybackslash}X
}
\midrule
\textbf{Category} & \textbf{Metric} & \textbf{Description} & \textbf{Evaluation} \\
\midrule
Retrieved literature
& Amount general
& How many sources does the tool provide?
& State the number of sources.
\\
\addlinespace

& Number of existing sources
& How many existing sources does the tool provide?
& State the number of existing sources.
\\
\addlinespace

& Number of scientific sources
& How many scientific sources does the tool provide?
& State the number of scientific sources.
\\
\addlinespace

& Number of peer-reviewed sources
& How many peer-reviewed sources does the tool provide?
& State the number of peer-reviewed sources.
\\
\addlinespace

& Intent match
& Does the source address the thematic scope of the user's information need?
& 
5 = source fits the thematic context of the user's need

4 =

3 = source partially fits the thematic context of the user's need

2 =

1 = source does not fit the thematic context of the user's need
\\
\addlinespace

& Number of sources with intent match
& How many sources with an intent match does the tool provide?
& State the number of sources with an intent match.
\\
\addlinespace

& Amount of usable sources
& How many usable sources for answering the research question does the tool provide?
& State the number of usable sources.

Usable sources = existing, scientific, peer-reviewed, and intent-matching sources; cut-off = minimum 3.
\\
\addlinespace

& Reproducibility
& Are the same or comparable results generated in different test trials?
& Comparison of sources found; value between 0 and 100\%.
\\
\bottomrule

\end{tabularx}
\end{table}
\newpage
\section{Q\&A Chat Prompts}
\begin{table}[htbp]
\caption{Q\&A single-document chat prompts}
\label{tab:single_doc_chat}
\centering
\small
\renewcommand{\arraystretch}{1.2}

\begin{tabularx}{\textwidth}{@{}p{0.28\textwidth}X@{}}
\toprule
\textbf{Metric} & \textbf{Prompt} \\
\midrule
Consistency external & How can the phenomenon of adaptive emotion regulation be defined in a broader sense? \\
 & From where and how were the study participants recruited? \\
 & The results of the PTSD symptom network analysis are consistent with the findings of which previous research? \\
\addlinespace
\addlinespace
No-answer criteria & Why do women develop PTSD more often than men? \\
\addlinespace
\addlinespace
Extraction (tables, formulas, graphs) & Extract the values from table 2. \\
 & Extract the formula for the edge between the nodes ``disturbing memories'' and ``disturbing dreams''. \\
 & Extract the values of the right plot in figure 1. \\
\addlinespace
Image information description & Describe the left image in figure 1. \\
\addlinespace
Single chat interpretation & Summarize the paper of Haws et al. (2022). \\
\bottomrule
\end{tabularx}
\end{table}
\newpage
\begin{table}[htbp]
\caption{Q\&A multi-document chat prompts}
\label{tab:multi_doc_chat}
\centering
\small
\renewcommand{\arraystretch}{1.2}
\begin{tabularx}{\textwidth}{@{}p{0.28\textwidth}X@{}}
\toprule
\textbf{Metric} & \textbf{Prompt} \\
\midrule
Consistency external & What is the facial feedback hypothesis? \\
 & Which previous findings support the results regarding the use of suppression as emotion regulation strategy? \\
 & Which facial muscles were measured in the articles using EMG? \\
\addlinespace
No-answer criteria & How does the Chatroom-Task work? \\
\addlinespace
Extraction (tables, formulas, graphs) & Extract the values from table 1 in Pizarro-Campagna et al. (2023). \\
 & Extract the formula, which states when the groups are significantly different at the .05 overall level. \\
 & Extract the scores from figure 2c in Matzke et al. (2014). \\
\addlinespace
Image information description & Describe figure 1 in Pizarro-Campagna et al. (2023). \\
\addlinespace
Multi chat interpretation & What are the similarities and differences between the participant samples examined in the papers? \\
 & What are similarities and differences between the measures used in the papers to measure emotional responses to the stimuli? \\
 & What are similarities and differences between the results of the papers Matzke et al. (2014) and Steinbrenner et al. (2022)? \\
\bottomrule
\end{tabularx}
\end{table}
\end{document}